\documentclass[runningheads]{llncs}
\usepackage[T1]{fontenc}
\usepackage{graphicx}
\usepackage{amsmath,amsfonts}
\usepackage{algorithmic}
\usepackage{algorithm}
\usepackage{array}
\usepackage{bigfoot}
\usepackage{textcomp}
\usepackage{stfloats}
\usepackage{float}
\usepackage{url}
\usepackage{verbatim}
\usepackage{booktabs}
\usepackage{bm}
\usepackage{pifont}
\usepackage{enumitem}
\usepackage{soul}
\usepackage{makecell}
\usepackage{tabularx}
\usepackage{multirow}
\usepackage[dvipsnames,table]{xcolor}
\usepackage{siunitx}
\usepackage{balance}
\usepackage{lscape}
\usepackage[utf8]{inputenc}
\usepackage{pmboxdraw}
\usepackage{placeins}

\usepackage{listings}
\definecolor{vscode-light-background}{RGB}{240,239,239}  
\definecolor{vscode-foreground}{RGB}{0,0,0}               
\definecolor{vscode-blue}{RGB}{56,92,170}                 
\definecolor{vscode-green}{RGB}{68,191,189}                
\definecolor{vscode-red}{RGB}{237,90,38}                  
\definecolor{vscode-purple}{RGB}{134,0,179}                
\definecolor{vscode-orange}{RGB}{214,103,0}                

\usepackage[capitalize]{cleveref}
\crefname{section}{Sec.}{Secs.}
\Crefname{section}{Section}{Sections}
\Crefname{table}{Table}{Tables}
\crefname{table}{Tab.}{Tabs.}

\def \eg {\emph{e.g.},}
\def \etal {\emph{et al.}}
\def \wrt {\emph{w.r.t.}}

\newcolumntype{Y}{>{\centering\arraybackslash}X}

\newcommand{\tit}[1]{\smallbreak\noindent\textbf{#1.}}

\begin{document}
\title{A Text Recognition Dataset from \\
Sahidic Coptic Ancient Manuscripts
}
%
%

\author{Fabio Quattrini\thanks{Equal contribution.}\orcidID{0009-0004-3244-6186} \and
Carmine Zaccagnino$^{\star,}$\orcidID{0009-0004-4953-6348} \and
Costanza Bianchi\orcidID{0009-0003-7522-7998} \and
Silvia Cascianelli\orcidID{0000-0001-7885-6050} \and
Rita Cucchiara\orcidID{0000-0002-2239-283X}}
\authorrunning{F. Quattrini et al.}
%

\institute{University of Modena and Reggio Emilia, Modena, Italy
\email{\{name.surname\}@unimore.it}
}
\maketitle              
\begin{abstract}
In this work, we target Handwritten Text Recognition (HTR) in low-resource scenarios, which arise from underrepresented languages, rare scripts, and degraded visual conditions typical of historical documents. We introduce SCAM (Sahidic Coptic Ancient Manuscripts), a new line-level dataset built from digitized ancient manuscripts written in the extinct Sahidic Coptic dialect. The dataset reflects a realistic and challenging setting, as it combines heterogeneous acquisition conditions across libraries with typical manuscript degradations such as ink fading, bleed-through, and material deterioration. In addition to visual complexity, SCAM poses significant linguistic challenges due to the scarcity of resources for Sahidic Coptic, its uncommon alphabet, and dialect-specific diacritics.
To support research in low-resource HTR, we benchmark several state-of-the-art approaches based on different paradigms, highlighting their limitations and strengths in this setting. Our results underline the gap between current HTR performance on well-resourced modern scripts and historically grounded, low-resource scenarios, thus providing a reference point for future developments. 
\end{abstract}
\section{Introduction}
\label{sec:intro}

\begin{figure}[t]
    \centering
        \includegraphics[height=20pt]{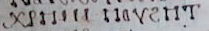}
        \includegraphics[height=20pt]{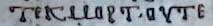}\\
        \includegraphics[height=20pt]{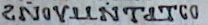}
        \includegraphics[height=20pt]{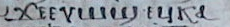}\\
        \includegraphics[height=20pt]{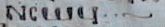}
        \includegraphics[height=20pt]{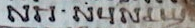}\\
        \includegraphics[height=30pt]{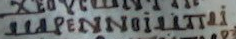}
        \includegraphics[height=30pt]{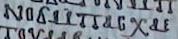}\\
        \includegraphics[height=30pt]{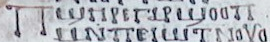}
        \includegraphics[height=40pt]{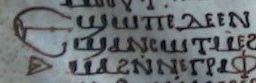}\\
        \vspace{0.2cm}
        \includegraphics[height=20pt]{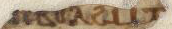}
        \includegraphics[height=20pt]{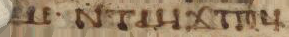}\\
        \includegraphics[height=20pt]{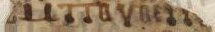}
        \includegraphics[height=20pt]{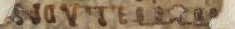}\\
        \includegraphics[height=20pt]{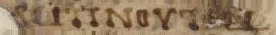}
        \includegraphics[height=20pt]{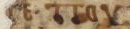}\\ 
        \includegraphics[height=30pt]{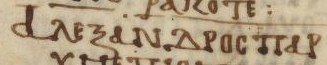}
        \includegraphics[height=30pt]{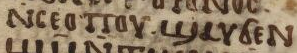}\\
        \includegraphics[height=30pt]{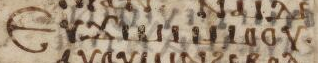}
        \includegraphics[height=30pt]{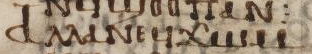}
    \vspace{-0.3cm}
    \caption{Sample images from the manuscript used to obtain the SCAM dataset. The top-four rows are from the SCAM-A subset, the bottom ones are from the SCAM-B.}\vspace{-0.6cm}
    \label{fig:coptic-samples}
\end{figure}

Handwritten Text Recognition (HTR) has made substantial progress in recent years, particularly due to the success of deep learning and end-to-end trainable architectures~\cite{alkendi2024advancements}. These systems, typically based on CNN-BiLSTM-CTC pipelines or Transformer encoders, achieve good recognition performance when trained on large, well-annotated datasets in modern languages and scripts. However, their performance significantly degrades when applied to low-resource scenarios, where data is limited, noisy, or involves rare scripts and underrepresented languages.
Such low-resource settings are common when dealing with historical manuscripts, poorly preserved documents, or extinct language scripts. In these contexts, the performance achievable with standard HTR pipelines often drop. 
To address this limitation, recent work has explored alternative strategies, such as synthetic data generation~\cite{kang2020ganwriting,bhunia2019handwriting}, and more recently, script-agnostic frameworks~\cite{baena2024general}. Nonetheless, some training is still required to achieve satisfactory performance, which realistically enables further usage of the transcribed documents by Humanities scholars.

In light of these considerations, we introduce a new line-level dataset featuring manuscripts in Sahidic Coptic, an underrepresented language of considerable interest to Coptologists and scholars of biblical studies, since it is among the languages into which the Bible was translated at an early stage. 
Our dataset is obtained from manuscript images preserved in different libraries and digitized with highly differing settings. This characteristic reflects a common fate of Coptic manuscripts from the White Monastery, many of which were dismantled in the 18th and 19th centuries and sold as individual leaves rather than as complete codices. 
From the point of view of the resulting images, the different acquisition settings and quality adds on the nuisances that are typical of ancient manuscripts (\eg~ink fading, bleed through, scratches, yellowed paper).
Moreover, the language of the manuscripts considered is another major challenge of the proposed dataset. In fact, Sahidic Coptic is no longer a spoken language, and the texts in the dataset are written in the Coptic script and employ a supralinear system characteristic of this language. 

To promote research in low-resource HTR, especially in the challenging scenario of manuscripts from different libraries, written in extinct languages, we devise our dataset, dubbed Sahidic Coptic Ancient Manuscript (SCAM)\footnote{More information and download link at \url{https://scam.silviacascianelli.com}}. The dataset features text lines extracted from pages preserved in two different archives and digitized under different conditions. Therefore, we organize the dataset into two subsets, dubbed SCAM-A and SCAM-B. Some example images used to build our dataset are reported in~\cref{fig:coptic-samples}.
In this paper, we analyze the visual and linguistic characteristics of our dataset, and benchmark different-paradigm State-of-the-Art HTR approaches, both to highlight the challenges of these data and the most promising solutions to face them, and to provide a leaderboard for future works.

\section{Related Work}
\label{sec:related}

HTR can be approached at various levels of granularity, including characters~\cite{jaderberg2015spatial,cilia2019ranking,clanuwat2019kuronet} (especially useful for ideogrammatic languages or rare scripts, such as ciphers), words~\cite{graves2009offline,bhunia2019handwriting,such2018fully,shi2016end}, lines~\cite{pham2014dropout,puigcerver2017multidimensional,voigtlaender2016handwriting,bluche2017gated,chowdhury2018efficient,moysset2017full,bluche2017scan}, paragraphs~\cite{bluche2017scan,bluche2016joint,wigington2018start}, or full pages~\cite{clanuwat2019kuronet,moysset2017full}. Among these, line-level HTR is the most widely adopted, typically assuming pre-segmented text lines~\cite{pham2014dropout,voigtlaender2016handwriting,bluche2017gated,puigcerver2017multidimensional,chowdhury2018efficient}. These systems can also be integrated with layout analysis and segmentation to achieve paragraph- or page-level recognition~\cite{bluche2017scan,moysset2017full,yousef2020origaminet}. In this work, we focus on pre-segmented line-level HTR, given its popularity and easy integration into page-level pipelines.

Early HTR systems were based on Hidden Markov Models (HMMs) to model sequences of explicitly-isolated character images and \emph{n}-gram language models to predict textual outputs~\cite{toselli2004integrated,toselli2015handwritten}. Such an explicit character isolation approach was then surpassed by end-to-end approaches, starting from the deep learning-based solution proposed by Graves~\etal~\cite{graves2009offline}, which consists of a multidimensional LSTM network that generates 2D feature maps directly from handwritten images, which are then reduced to 1D sequences and decoded using the Connectionist Temporal Classification (CTC) algorithm. This approach dominated the field~\cite{bluche2016joint,moysset20192d,de2019no} until computationally simpler and faster alternatives were proposed~\cite{shi2016end,puigcerver2017multidimensional}, using CNNs for feature extraction and 1D-LSTMs for sequence modeling. These backbones combining CNNs and recurrent models remain popular due to their strong performance and efficiency~\cite{pham2014dropout,voigtlaender2016handwriting,bluche2017gated,chowdhury2018efficient,cojocaru2020watch}. Nonetheless, 
a more recent direction models HTR as a sequence-to-sequence problem, where the image is represented as a sequence of patches and decoded into text using encoder-decoder architectures~\cite{sueiras2018offline,michael2019evaluating,zhang2019sequence,aberdam2021sequence}. Models implementing this paradigm can be trained by using cross-entropy loss, CTC loss, or both. Transformers~\cite{vaswani2017attention} have also been adopted for HTR to replace recurrent neural networks entirely, often leveraging large-scale pretraining on synthetic or real text data~\cite{kang2022pay,wick2021rescoring,li2021trocr}. 
Finally, it is worth mentioning that the emergence of Large Multimodal Models, such as GPT-4V~\cite{yang2023gpt4v}, has opened new avenues for HTR by enabling models to process and understand both visual and textual information. Preliminary explorations suggest that these models can perform HTR tasks with minimal task-specific supervision, although they currently lag behind specialized systems in accuracy and robustness, especially in low-resource settings.

\tit{HTR Datasets} 
Designing effective HTR systems requires large and diverse datasets that reflect the visual complexity of handwriting and the linguistic richness of the task. 
Widely used modern-language benchmarks include the IAM~\cite{marti2002iam} and RIMES~\cite{augustin2006rimes} datasets, which contain English and French lines, respectively. These datasets are written by multiple authors under controlled conditions. IAM writers transcribe sentences from the Lancaster-Oslo/Bergen corpus~\cite{johansson1978manual}, while RIMES contributors follow a script to simulate customer service communication. 

For historical HTR, the ICFHR14~\cite{sanchez2014icfhr2014}, ICFHR16~\cite{sanchez2016icfhr2016}, and ICFHR18~\cite{strauss2018icfhr2018} datasets are among the most notable. ICFHR14 includes English lines from the Bentham Papers~\cite{causer2012building}, mostly authored by Jeremy Bentham and collaborators. ICFHR16 (also called READ) focuses on ancient German manuscripts from the Ratsprotokolle collection, written over three centuries. ICFHR18 includes documents in German and Italian, from diverse time periods and collections, with test sets defined at the document level.
Other historical datasets include Rodrigo~\cite{serrano2010rodrigo} and Germana~\cite{perez2009germana}, which contain Spanish text written by single authors, and the LAM dataset~\cite{cascianelli2022lam}, which has been introduced to support research on ancient Italian single-author manuscripts with temporal variability in writing. 

Other small historical datasets exist, which are commonly adopted for exploring low-resource HTR scenarios, which is also the focus of this work. Examples of these datasets are George Washington~\cite{fischer2012lexicon}, Parzival~\cite{fischer2011transcription}, Saint Gall~\cite{fischer2009automatic}, Esposalles~\cite{romero2013esposalles},  Leopardi~\cite{cascianelli2021learning}, and~\cite{gillelevenson2023castilian}, which contain historical documents written by a few or single authors, often from narrow time spans, in a relatively well-represented language (English, Latin, and Italian). Particularly challenging examples of low-resource scenarios are explored via datasets of ciphered manuscripts, such as Borg and Copiale~\cite{baro2019towards}. These contain text in Latin and German, respectively, rendered in uncommon characters (either existing or abstract symbols). To further explore the low-resource HTR task, we propose a dataset of manuscript written in an underrepresented, extinct language using uncommon symbols. Specifically, we target ancient Sahidic Coptic, and devise a line-level dataset, dubbed SCAM, divided into two sections preserved in different places. 

For Coptic material, previous work has explored OCR for printed Coptic fonts through neural or multi-source training~\cite{miyagawa2019coptic,lincke2019coptic}. Existing digital resources, including CoptOT~\cite{coptot} and Coptic Scriptorium~\cite{coptic_scriptorium}, further provide important access points to Coptic textual material. While these platforms focus heavily on biblical text coverage or normalized linguistic corpora, they lack line-level layout annotations and diplomatic transcriptions. Consequently, SCAM fills this gap by introducing a dedicated, manuscript-specific HTR benchmark featuring cross-collection visual variations and intact line-level ground truth.

\section{Proposed Dataset}\label{sec:dataset}

Some recent works have targeted HTR in low-resource scenarios, also proposing datasets to better explore the task. By low-resource, one can refer to the visual appearance of the manuscript (given \eg~by its preservation state, or writer-specific handwriting, or both) or to the language of the manuscript, which is usually the case for historical documents in ancient languages. In this work, we devise a dataset for low-resource HTR that features all the challenges of this scenario. Specifically, our dataset, which we call SCAM, contains lines from 27 leaves, corresponding to 53 pages of an ancient manuscript written by a single author in Sahidic Coptic, an extinct language. Note that by ``leaf'' we mean a physical parchment sheet, which can accommodate one or two pages, written on both faces. The text is written in \textit{scriptio continua}, meaning that there are no spaces delimiting words.
The parchment manuscript, now known as Coptic Literary Manuscript (CLM) 359, is dated to 1002-1003 C.E., as indicated by the colophon, and comes from the White Monastery in Upper Egypt. This manuscript is of considerable importance, as it preserves substantial Coptic translations of Greek canonical and conciliar texts of the fourth and fifth centuries. It is not only a key witness to the development of Coptic canon law, but also an important source for reconstructing earlier stages of the Greek textual tradition. 
Note that SCAM differs from existing Coptic resources in both source material and benchmark design. \cite{coptot} mainly targets biblical witnesses and, although several records provide images and transcriptions, these materials are not systematically packaged as line-level HTR samples with polygon annotations and diplomatic line transcriptions. \cite{coptic_scriptorium} provides rich, normalized, and linguistically annotated Coptic corpora, but its editorial conventions often introduce word segmentation, normalization, punctuation, abbreviation expansion, or removal of supralinear marks, depending on the view and layer. By contrast, SCAM is built from one non-biblical Sahidic manuscript (a commentary on Church councils) whose surviving leaves are dispersed across institutions, preserving the manuscript's diplomatic line content and acquisition variability as the object of study. This makes SCAM complementary to previous Coptic resources: it is not intended to maximize textual coverage, but to isolate a realistic low-resource HTR setting with controlled scribe identity, cross-collection visual shift, and line-level ground truth.

To the best of our knowledge, only 49 leaves of the original manuscript have survived, which are now spread across countries, preserved in various libraries and cultural institutions~\footnote{\url{https://atlas.paths-erc.eu/manuscripts/359}}. 
As a result, each small set of leaves has been digitized under different conditions. This can significantly hinder the performance of an HTR model trained on samples from a certain collection when applied to samples in a different collection. To study this behavior, we collect samples from leaves preserved in two different libraries and organize the SCAM dataset into two subsets, dubbed SCAM-A and SCAM-B. 
The 15 leaves in the SCAM-A subset are digitized with professional scanners, under controlled lighting conditions, and with careful placement of the page in the scanner. On the other hand, the 12 leaves in the SCAM-B collection have been photographed with a relatively good-quality camera, without explicit control over the lighting conditions or the placement of the page.
In the following, we describe the process we followed for obtaining the samples from these pages, and analyze the visual and linguistic characteristics of the resulting dataset.

\subsection{Data Preparation} \label{subsec:data_preparation}
First, we isolate the line-level samples from all the page images contained in the collected leaves. Specifically, we resort to the popular open-source computer vision annotation tool CVAT\footnote{\url{https://www.cvat.ai/}} to trace polygons around each line. Note that we prefer using polygons instead of bounding boxes for two main reasons. The first is that some lines are not straight, either because the writer has laid them down without following a perfectly horizontal line or because of digitization artifacts. The second reason is more critical and specific of the manuscript at hand: since the manuscript is in scriptio continua, some of the lines start with bigger, ornate letters to indicate the beginning of a new paragraph or phrase. As a result, straight bounding boxes would lead to glyphs from other line samples be part of the line image of the sample with such slant or beginning letters (see~\Cref{fig:coptic-samples}). 

Once the lines have been isolated, an expert Coptologist provided their diplomatic transcriptions for each line, which are associated with the image by using CVAT. 
Note that, in this process, the Coptologist was able to exploit a special keyboard for Sahidic Coptic. The Sahidic Coptic characters are encoded in Unicode, but some of the graphemes are encoded using sequences of multiple code points, called \textbf{grapheme clusters}. This makes it possible to introduce variations of a character, such as accents. For model training and prediction, we ask the model to predict grapheme clusters and not individual code points, so that each glyph always corresponds to a single character. 

\subsection{Dataset Characteristics}
\label{sec:dataset_characteristics}
\tit{Statistics} 
Our dataset comprises 3240 lines, corresponding to 53 pages. It includes a total alphabet of 99 graphemes, with an average of 11.39 $\pm$ 2.04 characters per line. The data are partitioned into training, validation, and test splits as follows: the training set contains 2000 lines from 33 pages (89 unique graphemes), the validation set includes 610 lines from 10 pages (59 unique graphemes), and the test set consists of 630 lines from 10 pages (68 unique graphemes). 

As stated above, the images in our dataset are split into two subsets, SCAM-A and SCAM-B, depending on the archive in which the leaves are preserved. The characteristics of these subsets are the following:
\begin{itemize}[leftmargin=*, itemsep=0pt, topsep=0pt, parsep=0pt]
    \item \textbf{SCAM-A} contains 1445 lines, corresponding to 24 pages from 12 leaves. It features 87 graphemes, with an average line length of 11.52 $\pm$ 2.30 characters. The data are partitioned into training, validation, and test sets as follows: the training split includes 943 lines from 16 pages (76 unique graphemes), the validation split comprises 242 lines from 4 pages (48 unique graphemes), and the test split consists of 260 lines from 4 pages (61 unique graphemes).
    \item \textbf{SCAM-B} comprises 1795 lines, corresponding to 29 pages extracted from 15 leaves. It includes 67 distinct graphemes, with an average of 11.29 $\pm$ 1.81 characters per line. It is divided into training, validation, and test splits as follows: the training set contains 1057 lines from 17 pages (63 unique graphemes), the validation set includes 368 lines from 6 pages (51 unique graphemes), and the test set consists of 370 lines from 6 pages (50 unique graphemes).   
\end{itemize}
In~\Cref{tab:scam_stats} we quantify the visual and linguistic differences between the subsets. 

\begin{table}[t]
\centering
\setlength{\tabcolsep}{.8em}
\caption{Visual and linguistic differences between the two subsets (with RGB and grayscale images), expressed in terms of quantitative scores.\vspace{-0.3cm}}
\label{tab:scam_stats}
\resizebox{\columnwidth}{!}{%
\begin{tabular}{lc cccc} 
\toprule 
&& \textbf{FID} & \textbf{KID} &  \textbf{HWD} & \textbf{JSD}\\
\midrule
SCAM-A $\leftrightarrow$ SCAM-B (RGB) & & 113.62 & 0.13 & 1.06 & \multirow{2}{*}{0.49} \\
SCAM-A $\leftrightarrow$ SCAM-B (grayscale) & & ~65.06 & 0.07 & 1.04 & \\
\bottomrule
\end{tabular}
}
\end{table}

\tit{Visual Characteristics} First, we consider the visual characteristics, and quantify them in terms of Fréchet Inception Distance (\textbf{FID})~\cite{heusel2017gans} and the Kernel Inception Distance (\textbf{KID})~\cite{binkowski2018demystifying}. Note that these scores, commonly adopted in Image Generation evaluation, measure the dissimilarity between the distributions of visual features extracted from two sets of images, and therefore, we deem them suitable to compare the samples in the SCAM-A and SCAM-B subsets in terms of overall visual appearance. Moreover, we include in the analysis the Handwriting Distance (\textbf{HWD})~\cite{pippi2023hwd}, a recently-proposed score which isolates just the differences in writing style between pairs of images. The first row in the table contains the values of such scores computed on the original, RGB images (such as those in~\cref{fig:coptic-samples}). 
\begin{figure}[t]
\begin{minipage}[b]{0.230\linewidth}
\centering
    \includegraphics[width=\linewidth]{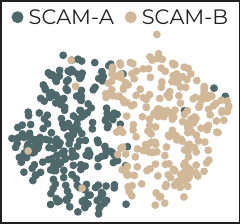}
    \vspace{-0.7cm}
\caption{t-SNE projection of VGG features extracted from RGB images in our dataset. 250 random points per subset plotted.} \label{fig:t-sne-rgb-slices}
\end{minipage}
\hfill
\begin{minipage}[b]{0.230\linewidth}
    \centering
    \includegraphics[width=\linewidth]{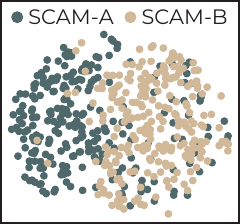}
    \vspace{-0.7cm}
    \caption{t-SNE projection of VGG features extracted from grayscale images in our dataset. 250 random points per subset plotted.}
    \label{fig:t-sne-gray-slices}
\end{minipage}
\hfill
\begin{minipage}[b]{0.230\linewidth}
    \centering
    \includegraphics[width=\linewidth]{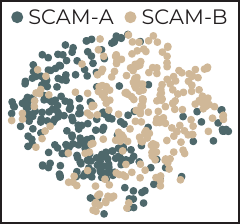}
    \vspace{-0.7cm}
    \caption{\strut t-SNE projection of HWD features extracted from RGB images in our dataset. 250 random points per subset plotted.}
    \label{fig:t-sne-rgb-hwd-slices}
\end{minipage}
\hfill
\begin{minipage}[b]{0.230\linewidth}
    \centering
    \includegraphics[width=\linewidth]{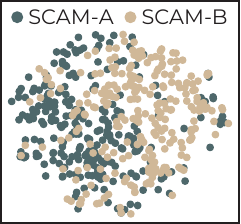}
    \vspace{-0.7cm}
    \caption{\strut t-SNE projection of HWD features extracted from grayscale images in our dataset. 250 random points per subset plotted.}
    \label{fig:t-sne-gray-hwd-slices}
\end{minipage}
\vspace{-.50cm}
\end{figure}
It emerges that the perceptual features are quite distinct and that the writing style in the two subsets is related but not identical. In particular, as shown by~\cite{pippi2023hwd}, typical values for HWD in same-author sets is between 0 and 1.5. This reflects the fact that the manuscript from which SCAM was derived was written by the same person. We deepen our analysis by applying a grayscale conversion to the images of both subsets, as this is a simple transformation that practitioners can easily adopt with the aim of reducing the visual gap between the subsets. Alternatively, document restoration approaches such as binarization~\cite{otsu1979threshold,quattrini2024fourbi} could be applied, but once sufficiently explored for our data. In fact, blindly applying such approaches can result in text content alteration in the lines. As it can be observed from the values in the second row of~\Cref{tab:scam_stats}, the difference in visual features is reduced but not entirely eliminated, underlining that the SCAM dataset poses visual challenges that are not trivial to overcome. 

These aspects are even clearer when looking at t-SNE projections of visual features extracted from both subsets (\cref{fig:t-sne-rgb-slices,fig:t-sne-rgb-hwd-slices,fig:t-sne-gray-slices,fig:t-sne-gray-hwd-slices}). In particular, we project features extracted by a VGG trained on natural images and by the VGG backbone of the HWD score, trained to distinguish handwriting styles. We argue that the first feature set captures the overall appearance (\eg~color and texture) of the images, while the second set captures the handwriting style characteristics (\eg~thickness, slant, and kerning). Note that both feature sets are extracted from the original RGB images in the subsets and from their grayscale version. From~\cref{fig:t-sne-rgb-slices}, we can see that the VGG features on RGB images are clearly divided into two separate clusters, which get closer when computed on grayscale images, as shown in~\Cref{fig:t-sne-gray-slices}. The HWD features, as shown for RGB and grayscale images in~\Cref{fig:t-sne-rgb-hwd-slices} and in~\Cref{fig:t-sne-gray-hwd-slices} respectively, are less separated. This suggests that, although the handwriting in the two subsets is the same (the images come from a single-author manuscript), the differences between the images in the two subsets in terms of preservation state and digitization settings have also impacted the characteristics of the handwriting (\eg~by altering the thickness).

\begin{figure}[t]
    \centering
    \includegraphics[width=\linewidth]{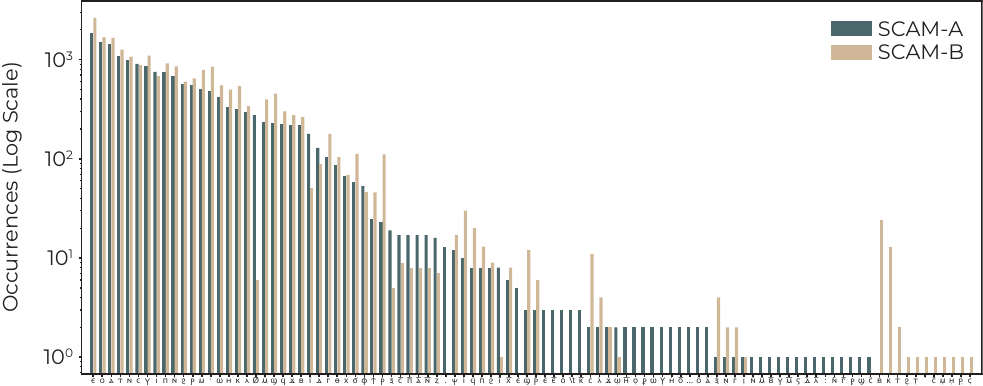}
    \vspace{-.7cm}
    \caption{Graphemes distribution in the two subsets of SCAM, in logarithmic scale. For comparison, graphemes are in descending order of frequency as computed in SCAM-A.}
    \label{fig:char_distribution}
\vspace{-.3cm}
\end{figure}

\tit{Linguistic Characteristics}
In~\Cref{tab:scam_stats} we also report the Jensen-Shannon divergence (\textbf{JSD}) computed on $n$-grams (with $n = 1, ..., 5$) extracted from the textual transcriptions of the SCAM-A and SCAM-B subsets. This score is considered to obtain a measure of the similarity of the text content of the two subsets. The value of the score suggests that, even though the dataset contains text taken from the same manuscript treating the same topic, there are non-negligible differences in the text content. 

We deepen this analysis by plotting the distribution of the graphemes on the two subsets (in~\cref{fig:char_distribution}). It can be observed that, as happens in languages, the most common characters are equally frequent in both subsets. However, both subsets contain unique graphemes that are not present in the other, with SCAM-A having more of such cases (31 graphemes that are not present in SCAM-B, versus 11 graphemes in SCAM-B that are not in SCAM-A).

\section{Experimental Analysis}
\label{sec:experiments}

This section presents an experimental analysis of the performance of several state-of-the-art models. The evaluation is performed by training each of them on \textbf{SCAM-A}, \textbf{SCAM-B} and the full \textbf{SCAM} dataset. Then, each model variant is tested on both the SCAM-A and SCAM-B subsets separately and jointly. 

\tit{Considered Scores}
To assess performance, we report the Character Error Rate (\textbf{CER}) and the Sequence Error Rate (\textbf{SER}). Following standard conventions in Handwritten Text Recognition (HTR), these scores are derived from the Edit Distance between the predicted text and the corresponding ground truth. 
Specifically, we compute the minimum number of substitutions, deletions, and insertions required to transform the predicted sequence into the ground truth. To determine the CER and SER for the entire test set, we sum the distances of all samples and normalize by the total number of units in the ground truth. For the CER, the distance is computed at the character level, while for the SER, it is computed on the content of the entire line sample, since our dataset is in scriptio continua.

\tit{Considered Approaches}
We consider a range of model architectures and sizes in order to showcase the challenging nature of this dataset, and highlight which models perform best on it. In particular, we test the following architectures:
\begin{itemize}[leftmargin=*, itemsep=0pt, topsep=0pt, parsep=0pt]
    \item We include the model proposed in~\cite{puigcerver2017multidimensional}, dubbed \textbf{CRNN}, which combines a convolutional backbone with BLSTM layers and is trained end-to-end to optimize the CTC loss. This model, which represents one of the most popular paradigms for learning-based HTR, serves as our primary recurrent baseline for sequence modeling without explicit alignment.
    \item We also consider the Sequence-to-Sequence recurrent attention paradigm by including the approach proposed by \textbf{Michael et al.}~\cite{michael2019evaluating}. This approach combines a CNN-BiLSTM encoder with an attention-based LSTM decoder and auxiliary CTC loss, serving as a strong RNN-based attention baseline.
    \item Additionally, we consider models that rely on the attention operation instead of recurrent components, namely the Vertical Attention Network (\textbf{VAN}) proposed in~\cite{coquenet2023endtoend} and the Convolutional Self-Attention Network (\textbf{C-SAN}) from~\cite{arce2022self}. These models replace RNNs with gated or self-attention mechanisms to enable parallelization and more efficient training, assessing whether attention alone can model handwriting dependencies.
    \item Another strategy to replace the recurrent component of the HTR pipeline entails exploiting a Transformer encoder-decoder, as \textbf{LT}~\cite{light_barrere_2022}, \textbf{VLT}~\cite{training_barrere_2024}, and the one proposed by \textbf{Kang et al.}~\cite{kang2022pay}, or a Transformer encoder, as done in \textbf{HTR-VT}~\cite{li_htr-vt_2025}. In both cases, the visual features are extracted by a convolutional backbone, and the Transformer-based decoder improves parallelization and captures long-range dependencies in handwriting.
    \item Finally, we consider \textbf{TrOCR}~\cite{li2021trocr}, which represents the fully Transformer-based, end-to-end paradigm. Specifically, the image features are extracted with a Transformer encoder, and the transcription is obtained with a Transformer decoder. Both components have been pre-trained on synthetic data.
\end{itemize}
\tit{Implementation Details} For the experimental evaluation, we finetune TrOCR using the official HuggingFace implementation~\cite{wolf-etal-2020-transformers}. For each model size, we initialize the weights from the respective \textit{handwritten} released checkpoint\footnote{\texttt{hf.co/microsoft/trocr-\textit{size}-handwritten}, where size $\in$ \{small, base, large\}}.  
For the other architectures, we base our code on the HTR models implementations available online\footnote{\url{https://github.com/carlos10garrido/HTR-OOD/}} and train them from scratch. To this end, we modify the tokenizer, removing the ASCII transliteration step\footnote{\url{https://pypi.org/project/Unidecode/}}, due to its incompatibility with non-Latin scripts and, therefore, our graphemes (\Cref{subsec:data_preparation}).
Each model is trained (or fine-tuned, in the case of the TrOCR variants) with an early stopping strategy based on the CER computed on the validation set of the dataset used for training, with patience set to 100 epochs.
    
\subsection{Results}
\begin{figure*}[t]
    \centering
    \includegraphics[width=\linewidth]{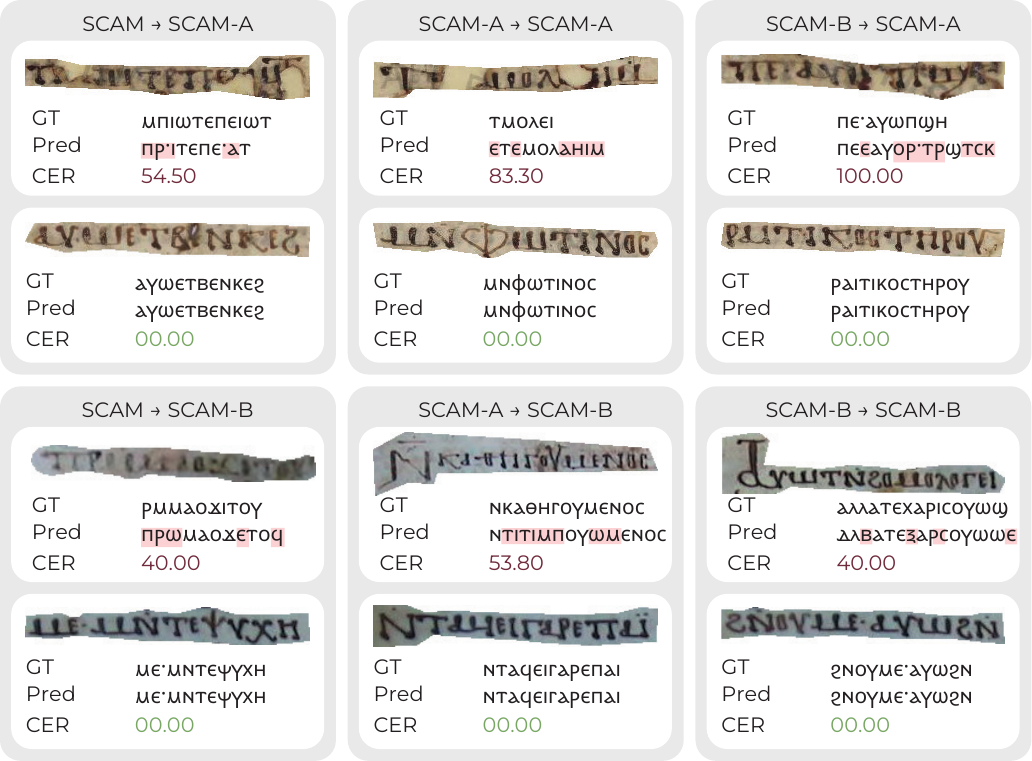}
    \vspace{-0.7cm}
    \caption{Qualitative results on the considered setups (Train dataset $\rightarrow$ Test dataset) of the overall best-performing benchmarked model (VAN). We also report the CER on each considered sample and highlight the errors in red.}
    \label{fig:qualitatives}
    \vspace{-0.5cm}
\end{figure*}

In the following, we report an experimental analysis, both quantitative (\Cref{tab:intra_dataset,tab:inter_dataset,tab:inter_dataset_gray,tab:full_dataset}) and qualitative (\cref{fig:qualitatives}) of the considered approaches, and discuss the insights drawn from such analysis.

\tit{Architecture type matters more than scale for rare scripts}
Across all tables, we can observe that VAN achieves the best CER and SER on both subsets (some examples of its best and worst transcriptions over the different setups are reported in~\cref{fig:qualitatives}). We argue that the gated vertical-attention mechanism is particularly well-suited to the regular layout of the Coptic script glyphs: by attending to vertical strips of the image, VAN naturally aligns with the character-level structure of the lines without relying on explicit language priors. In contrast, TrOCR variants, despite their large-scale synthetic pre-training on Latin-script handwriting, do not achieve competitive results. These models rely on an autoregressive language prior at decoding time, while CTC-based models condition each predicted frame based on visual features independently. This pattern suggests that Transformer decoders pre-trained on high-resource scripts cannot efficiently adapt their learned language priors to an unseen script under data-scarce fine-tuning, and that model scale is not a substitute for architectural alignment with the task. Additionally, the scriptio continua writing provides a particularly challenging setting, which prevents the model from easily learning word priors.
We also note that SER values are consistently high across all models and settings, even when the CER is relatively low. This is a direct consequence of the scriptio continua: with an average line length of $11.39 \pm 2.04$ characters, even a model achieving $\sim$7.5\ CER is expected to produce at least one error in roughly half of all lines, counted as a full-line mismatch in the SER. 

\begin{table}[t]
\centering
\setlength{\tabcolsep}{0.8em}
\caption{Performance obtained by models trained on the entire SCAM training set and tested on the whole test set (SCAM) and on each test subset (SCAM-A and SCAM-B).\vspace{-0.3cm}}
\label{tab:full_dataset}
\resizebox{\columnwidth}{!}{%
\begin{tabular}{lc cc c cc c cc}
\toprule 
&& \multicolumn{2}{c}{\textbf{SCAM}} 
&& \multicolumn{2}{c}{\textbf{SCAM-A}}
&& \multicolumn{2}{c}{\textbf{SCAM-B}}\\
\cmidrule{3-4} \cmidrule{6-7} \cmidrule{9-10}
&& \textbf{CER} & \textbf{SER} & & \textbf{CER} & \textbf{SER} & & \textbf{CER} & \textbf{SER}\\
\midrule
\textbf{CRNN~\cite{puigcerver2017multidimensional}} & & 9.47 & 49.79 & & 9.00 & 51.54 & & 9.85 & 48.38 \\
\textbf{C-SAN~\cite{arce2022self}}                  & & 88.06 & 100.00 & & 87.52 & 100.00 & & 88.50 & 100.00 \\
\textbf{VAN~\cite{coquenet2023endtoend}}            & & 5.25 & 31.99 & & 5.96 & 36.15 & & 4.68 & 28.65 \\
\textbf{HTR-VIT~\cite{li_htr-vt_2025}}              & & 10.27 & 59.62 & & 10.98 & 66.54 & & 9.70 & 54.05 \\
\textbf{Kang et al.~\cite{kang2022pay}}             & & 11.43 & 54.18 & & 11.44 & 57.69 & & 11.42 & 51.35 \\
\textbf{Michael et al.~\cite{michael2019evaluating}} & & 33.35 & 91.16 & & 36.19 & 95.39 & & 31.07 & 87.76 \\
\textbf{LT~\cite{light_barrere_2022}}               & & 22.59 & 78.74 & & 21.86 & 78.85 & & 23.18 & 78.65 \\ 
\textbf{VLT~\cite{training_barrere_2024}}           & & 15.03 & 62.81 & & 14.89 & 67.31 & & 15.15 & 59.19 \\
\textbf{TrOCR-S~\cite{li2021trocr}}                 & & 13.12 & 65.99 & & 17.04 & 86.87 & & 9.96 & 49.20 \\
\textbf{TrOCR-B~\cite{li2021trocr}}                 & & 12.42 & 74.39 & & 14.42 & 77.78 & & 10.81 & 71.66 \\
\textbf{TrOCR-L~\cite{li2021trocr}}                 & & 11.33 & 72.63 & & 12.82 & 74.50 & & 10.14 & 71.12 \\
\bottomrule
\end{tabular}
}
\end{table}

\begin{table}
\centering
\setlength{\tabcolsep}{1.4em}
\caption{Performance obtained by models trained and tested on the same subset. ``A'' indicates the SCAM-A subset, ``B'' the SCAM-B.\vspace{-0.3cm}}
\label{tab:intra_dataset}
\resizebox{\columnwidth}{!}{%
\begin{tabular}{lc cc c cc}
\toprule 
&& \multicolumn{2}{c}{\textbf{A$\rightarrow$A}} 
&& \multicolumn{2}{c}{\textbf{B$\rightarrow$B}}\\
\cmidrule{3-4} \cmidrule{6-7}
&& \textbf{CER} & \textbf{SER} & & \textbf{CER} & \textbf{SER}\\
\midrule
\textbf{CRNN~\cite{puigcerver2017multidimensional}} & & 13.91 & 65.39 & & 14.63 & 64.32 \\  
\textbf{C-SAN~\cite{arce2022self}                 } & & 87.52 & 100.00 & & 9.87 & 44.05 \\ 
\textbf{VAN~\cite{coquenet2023endtoend}           } & & 7.50 & 40.77 & & 7.58 & 41.08 \\ 
\textbf{HTR-VIT~\cite{li_htr-vt_2025}             } & & 15.24 & 74.23 & & 14.48 & 71.08 \\  
\textbf{Kang et al.~\cite{kang2022pay}                 } & & 14.16 & 65.39 & & 18.38 & 67.84 \\  
\textbf{Michael et al.~\cite{michael2019evaluating}              } & & 79.15 & 98.46 & & 82.00 & 100.00 \\ 
\textbf{LT~\cite{light_barrere_2022}                          } & & 44.28 & 98.46 & & 46.99 & 97.30 \\ 
\textbf{VLT~\cite{training_barrere_2024}          } & & 21.48 & 80.00 & & 32.94 & 88.92 \\
\textbf{TrOCR-S~\cite{li2021trocr}                } & & 30.70 & 98.23 & & 12.11 & 56.15 \\
\textbf{TrOCR-B~\cite{li2021trocr}                } & & 18.57 & 81.57 & & 12.36 & 75.40 \\
\textbf{TrOCR-L~\cite{li2021trocr}                } & & 14.99 & 78.54 & & 10.59 & 68.45 \\
\bottomrule
\end{tabular}
}\vspace{-.3cm}
\end{table}

\tit{Data breadth matters more than model scale}
In~\Cref{tab:full_dataset}, we report results when training on the full SCAM dataset and testing on each subset. This represents the most favorable scenario, with models trained on annotated samples from multiple archives. 
Training on SCAM brings substantial improvements with respect to training on just one subset for most models. 
We argue that these results confirm that the in-domain failures observed in~\Cref{tab:intra_dataset} are primarily due to data scarcity rather than to a fundamental incompatibility with the Sahidic Coptic script. Overall, these results show that using in-domain training data is more important than model capacity, as lightweight VAN and CRNN outperform the pre-trained TrOCR-L.    

\begin{table}[t]
\centering
\setlength{\tabcolsep}{1.4em}
\caption{Performance obtained by models trained on a subset and tested on a different subset. ``A'' indicates the SCAM-A subset, ``B'' the SCAM-B.\vspace{-0.3cm}}
\label{tab:inter_dataset}
\resizebox{\columnwidth}{!}{%
\begin{tabular}{lc cc c cc}
\toprule 
&& \multicolumn{2}{c}{\textbf{A$\rightarrow$B}} 
&& \multicolumn{2}{c}{\textbf{B$\rightarrow$A}}\\
\cmidrule{3-4} \cmidrule{6-7}
&& \textbf{CER} & \textbf{SER} & & \textbf{CER} & \textbf{SER}\\
\midrule
\textbf{CRNN~\cite{puigcerver2017multidimensional}} & & 26.63 & 76.76 & & 31.28 & 93.85 \\
\textbf{C-SAN~\cite{arce2022self}}                  & & 88.50 & 100.00 & & 23.54 & 88.46 \\
\textbf{VAN~\cite{coquenet2023endtoend}}            & & 14.44 & 55.41 & & 19.94 & 79.61 \\
\textbf{HTR-VIT~\cite{li_htr-vt_2025}}              & & 29.76 & 89.46 & & 25.31 & 94.23 \\
\textbf{Kang et al.~\cite{kang2022pay}}             & & 11.44 & 57.69 & & 32.92 & 93.08 \\
\textbf{Michael et al.~\cite{michael2019evaluating}}& & 81.48 & 100.00 & & 80.51 & 99.62 \\
\textbf{LT~\cite{light_barrere_2022}}               & & 64.03 & 99.73 & & 62.03 & 99.62 \\
\textbf{VLT~\cite{training_barrere_2024}}           & & 37.50 & 88.92 & & 47.28 & 98.85 \\
\textbf{TrOCR-S~\cite{li2021trocr}}                 & & 45.46 & 98.13 & & 30.37 & 98.49 \\
\textbf{TrOCR-B~\cite{li2021trocr}}                 & & 24.18 & 92.25 & & 18.36 & 90.15 \\
\textbf{TrOCR-L~\cite{li2021trocr}}                 & & 14.15 & 79.95 & & 18.36 & 90.15 \\
\bottomrule
\end{tabular}
}
\end{table}

\begin{table}
\centering
\setlength{\tabcolsep}{1.4em}
\caption{Performance obtained by models trained on a subset and tested on a different subset, both in grayscale. ``A'' indicates the SCAM-A subset, ``B'' the SCAM-B.\vspace{-0.3cm}}
\label{tab:inter_dataset_gray}
\resizebox{\columnwidth}{!}{%
\begin{tabular}{lc cc c cc}
\toprule 
&& \multicolumn{2}{c}{\textbf{A$\rightarrow$B}} 
&& \multicolumn{2}{c}{\textbf{B$\rightarrow$A}}\\
\cmidrule{3-4} \cmidrule{6-7}
&& \textbf{CER} & \textbf{SER} & & \textbf{CER} & \textbf{SER}\\
\midrule
\textbf{CRNN~\cite{puigcerver2017multidimensional}} & & 26.22 & 78.11 & & 28.63 & 89.23 \\ 
\textbf{C-SAN~\cite{arce2022self}}                  & & 87.50 & 100.00 & & 85.36 & 100.00 \\
\textbf{VAN~\cite{coquenet2023endtoend}}            & & 12.69 & 52.93  & & 17.96 & 77.31  \\ 
\textbf{HTR-VIT~\cite{li_htr-vt_2025}}              & & 27.65 & 85.68  & & 25.52 & 90.77  \\  
\textbf{Kang et al.~\cite{kang2022pay}}             & & 26.91 & 81.08  & & 35.91 & 96.54  \\ 
\textbf{Michael et al.~\cite{michael2019evaluating}}& & 80.81 & 100.00 & & 81.35 & 100.00 \\
\textbf{LT~\cite{light_barrere_2022}}               & & 63.29 & 98.92  & & 60.01 & 99.62  \\ 
\textbf{VLT~\cite{training_barrere_2024}}           & & 34.13 & 88.65  & & 51.74 & 99.62  \\ 
\textbf{TrOCR-S~\cite{li2021trocr}}                 & & 22.73 & 92.42  & & 25.38 & 84.23  \\
\textbf{TrOCR-B~\cite{li2021trocr}}                 & & 17.73 & 82.58  & & 18.98 & 89.31  \\
\textbf{TrOCR-L~\cite{li2021trocr}}                 & & 14.84 & 76.52  & & 15.77 & 83.70  \\
\bottomrule
\end{tabular}
}\vspace{-0.2cm}
\end{table}
\tit{Domain generalization across subsets}
In~\Cref{tab:inter_dataset}, we evaluate the performance of models trained on one subset and tested on the other. This setting directly mirrors a real-world scenario, where a scholar might annotate leaves from one library, train a model, and then apply it to newly-digitized leaves from a different archive. Note that a scholar does not always have physical access to the archives and, therefore, control over digitization conditions. New leaves might have different acquisition conditions and contain graphemes absent from the training data (\eg~SCAM-A and SCAM-B have only partially overlapping grapheme vocabularies of 87 and 67 symbols, respectively, as shown in~\Cref{fig:char_distribution}). This is an interesting setting that can be explored with the SCAM dataset. We can observe that training on SCAM-B and testing on SCAM-A is consistently harder than doing the opposite for almost every model. We attribute this to the lower grapheme count of SCAM-B. Training on the SCAM-A, which has broader glyph coverage and cleaner images, provides a richer and more generalizable foundation. 
This asymmetry shows that high-quality library scans with broad grapheme coverage lead to data that can be used to train models that generalize better to field photographs than vice versa. This result offers a guiding principle for practitioners planning transcription pipelines, and defines a challenge for scholars who want to tackle domain generalization in scenarios similar to the one in our dataset

\tit{Image preprocessing helps reduce domain gap}
In~\Cref{tab:inter_dataset_gray} we evaluate the effect of a simple image preprocessing step, namely conversion to grayscale, on the inter-subset performance. From the results, we can see that this step consistently improves performance across most models. As a notable example, we observe that, in this setting, TrOCR-L reaches comparable performance \wrt VAN in CER values, with slightly worse SER values, revealing that bridging the color gap is not sufficient to achieve reliable line-level transcription. This is consistent with the dataset statistics (\Cref{tab:scam_stats}): converting to grayscale nearly halves the FID, but the HWD score barely changes, indicating a residual visual gap. Furthermore, in a scriptio continua text, a single character error in a long boundary-free line leads to a full-line SER mismatch. 

\tit{Hybrid decoders underperform CTC-only approaches}
A consistent finding across all three experimental settings (intra-subset - \Cref{tab:intra_dataset}, inter-subset - \Cref{tab:inter_dataset}, full-dataset - \Cref{tab:full_dataset}) is the unsatisfactory performance of the three \textit{hybrid} architectures (Michael et al., LT, and VLT), which combine a CTC branch with an additional attention or cross-entropy decoder, compared to approaches relying solely on CTC decoding. From the results, we can infer that the joint CTC+decoder training objective does not converge well on the proposed SCAM dataset. 
Pure CTC-based models (CRNN, VAN) do not rely on an explicit autoregressive prior at decoding time and are agnostic to word delimiters, making them the most reliable paradigm in this low-resource setting. 
\section{Conclusion}
\label{sec:conclusion}

In this paper, we have introduced SCAM, a new line-level dataset in the underrepresented Sahidic Coptic language, derived from the 11\textsuperscript{th}-century CLM 359 manuscript and manually annotated by domain experts. By gathering sources from different libraries and digitized under heterogeneous acquisition conditions, we have modeled a realistic and challenging low-resource scenario featuring visual variability, degradation phenomena, and linguistic complexity.

Our experimental evaluation of several state-of-the-art HTR architectures across different sub-splits reveals several findings of broader relevance for Document AI. First, pure CTC-based architectures proved to be the most robust choice for scriptio continua manuscripts in resource-scarce languages: by not relying on an explicit autoregressive decoder, they are less prone to error cascades and scale better with limited data than hybrid models, which combine CTC with attention or cross-entropy decoders. Second, the domain gap between SCAM's two subsets results in models trained on SCAM-B consistently generalizing worse to SCAM-A than vice versa. Third, grayscale preprocessing helps narrow the cross-archive gap when multi-domain data is unavailable but is not a substitute for training on diverse, multi-archive data. When such data is available, models like VAN provide the best accuracy across all settings. When only a single archive is available, simpler architectures like CRNN with grayscale preprocessing can combine robustness and adaptability to cross-domain deployment.

More broadly, these findings underscore the value of SCAM's dual-archive structure, exposing the domain asymmetry, the grapheme coverage gap, or the architectural sensitivity of certain models to specific visual distributions. We release SCAM as a benchmark to support future research on low-resource, rare-script, and cross-domain handwritten text recognition, and we hope it will encourage the development of HTR methods better suited to the challenges posed by real historical manuscript collections.

\begin{credits}
\subsubsection{\ackname} 
This work was supported by the PNRR project Italian Strengthening of ESFRI RI Resilience (ITSERR) funded by the European Union – NextGenerationEU, the ``AI for Digital Humanities'' project funded by ``Fondazione di Modena'', the GCP Credit Award, the Google Cloud Research Credits program with the award GCP19980904, and the FARD2025 project (CUP E93C25000370005).
We acknowledge EuroHPC Joint Undertaking and ISCRA for awarding us access to LUMI at CSC, Finland, LEONARDO at CINECA, Italy, and MareNostrum5 at BSC, Spain. 

\subsubsection{\discintname}
The authors have no competing interests to declare that are relevant to the content of this article.
\end{credits}

%
%
\bibliographystyle{splncs04}
\bibliography{main}

@string{nips      = {Advances in Neural Information Processing Systems}}

@string{eccv      = {Proceedings of the European Conference on Computer Vision}}

@string{cvpr      = {Proceedings of the IEEE/CVF Conference on Computer Vision and Pattern Recognition}}

@string{iclr      = {Proceedings of the International Conference on Learning Representations}}

@string{bmvc      = {Proceedings of the British Machine Vision Conference}}

@string{aaai      = {Proceedings of the AAAI Conference on Artificial Intelligence}}

@string{emnlp     = {Proceedings of the Conference on Empirical Methods in Natural Language Processing}}

@string{icpr      = {Proceedings of the International Conference on Pattern Recognition}}

@string{icdar     = {Proceedings of the International Conference on Document Analysis and Recognition}}

@string{icfhr     = {Proceedings of the International Conference on Frontiers in Handwriting Recognition}}

@string{iwfhr     = {Proceedings of the International Workshop on Frontiers in Handwriting Recognition}}

@string{hip       = {Proceedings of the International Workshop on Historical Document Imaging and Processing}}

@string{lrec      = {Proceedings of the Language Resources and Evaluation Conference}}

@string{das       = {Proceedings of the International Workshop on Document Analysis Systems}}

@string{ieeetpami = {IEEE Transactions on Pattern Analysis and Machine Intelligence}}

@string{ieeetsmc  = {IEEE Transactions on Systems, Man, and Cybernetics}}

@string{pr        = {Pattern Recognit.}}

@string{prl       = {Pattern Recognit. Lett.}}

@string{ijdar     = {International Journal on Document Analysis and Recognition}}

@string{ijprai    = {International Journal of Pattern Recognition and Artificial Intelligence}}

@string{nips     = {NeurIPS}}

@string{cvpr     = {CVPR}}

@string{iclr     = {ICLR}}

@string{bmvc     = {BMVC}}

@string{eccv     = {ECCV}}

@string{icpr     = {ICPR}}

@string{aaai     = {AAAI}}

@string{emnlp    = {EMNLP}}

@string{ijdar    = {IJDAR}}

@string{icdar    = {ICDAR}}

@string{icfhr    = {ICFHR}}

@string{iwfhr    = {IWFHR}}

@string{hip      = {HIP}}

@string{lrec     = {LREC}}

@string{das      = {DAS}}

@string{ieeetpami  = {IEEE Trans. PAMI}}

@string{ijprai     = {Int. J. Pattern Recognit. Artif. Intell.}}

@book{johansson1978manual,
	title        = {{Manual of information to accompany the Lancaster-Oslo/Bergen Corpus of British English, for use with digital computer}},
	author       = {Johansson, Stig and Leech, Geoffrey N and Goodluck, Helen},
	year         = {1978},
	publisher    = {Department of English, University of Oslo}
}

@article{marti2002iam,
	title        = {{The IAM-database: an English sentence database for offline handwriting recognition}},
	author       = {Marti, U-V and Bunke, Horst},
	year         = {2002},
	journal      = ijdar,
	volume       = {5},
	number       = {1},
	pages        = {39--46}
}

@article{toselli2004integrated,
	title        = {Integrated handwriting recognition and interpretation using finite-state models},
	author       = {Toselli, Alejandro H and Juan, Alfons and Gonz{\'a}lez, Jorge and Salvador, Ismael and Vidal, Enrique and Casacuberta, Francisco and Keysers, Daniel and Ney, Hermann},
	year         = {2004},
	journal      = ijprai,
	volume       = {18},
	number       = {04},
	pages        = {519--539}
}

@inproceedings{augustin2006rimes,
	title        = {{RIMES evaluation campaign for handwritten mail processing}},
	author       = {Augustin, Emmanuel and Carr{\'e}, Matthieu and Grosicki, Emmanu{\`e}le and Brodin, J-M and Geoffrois, Edouard and Pr{\^e}teux, Fran{\c{c}}oise},
	year         = {2006},
	booktitle    = iwfhr,
	pages        = {231--235}
}

@inproceedings{graves2009offline,
	title        = {{Offline Handwriting Recognition with Multidimensional Recurrent Neural Networks}},
	author       = {Graves, Alex and Schmidhuber, J{\"u}rgen},
	year         = {2008},
	booktitle    = nips
}

@inproceedings{fischer2009automatic,
	title        = {Automatic transcription of handwritten medieval documents},
	author       = {Fischer, Andreas and Wuthrich, Markus and Liwicki, Marcus and Frinken, Volkmar and Bunke, Horst and Viehhauser, Gabriel and Stolz, Michael},
	year         = {2009},
	booktitle    = {VSMM}
}

@inproceedings{serrano2010rodrigo,
	title        = {{The RODRIGO Database}},
	author       = {Serrano, Nicol{\'a}s and Castro, Francisco and Juan, Alfons},
	year         = {2010},
	booktitle    = lrec
}

@inproceedings{perez2009germana,
	title        = {{The GERMANA Database}},
	author       = {P{\'e}rez, Daniel and Taraz{\'o}n, Lionel and Serrano, Nicol{\'a}s and Castro, Francisco and Ramos Terrades, Oriol and Juan, Alfons},
	year         = {2010},
	booktitle    = lrec
}

@inproceedings{fischer2011transcription,
	title        = {{Transcription alignment of Latin manuscripts using hidden Markov models}},
	author       = {Fischer, Andreas and Frinken, Volkmar and Forn{\'e}s, Alicia and Bunke, Horst},
	year         = {2011},
	booktitle    = hip
}

@article{fischer2012lexicon,
	title        = {{Lexicon-free handwritten word spotting using character HMMs}},
	author       = {Fischer, Andreas and Keller, Andreas and Frinken, Volkmar and Bunke, Horst},
	year         = {2012},
	journal      = prl,
	publisher    = {Elsevier},
	volume       = {33},
	number       = {7},
	pages        = {934--942}
}

@article{causer2012building,
	title        = {Building a volunteer community: results and findings from Transcribe Bentham},
	author       = {Causer, Tim and Wallace, Valerie},
	year         = {2012},
	journal      = {Digit. Humanit. Q.},
	volume       = {6},
	number       = {2}
}

@article{romero2013esposalles,
	title        = {{The ESPOSALLES database: An ancient marriage license corpus for off-line handwriting recognition}},
	author       = {Romero, Ver{\'o}nica and Forn{\'e}s, Alicia and Serrano, Nicol{\'a}s and S{\'a}nchez, Joan Andreu and Toselli, Alejandro H and Frinken, Volkmar and Vidal, Enrique and Llad{\'o}s, Josep},
	year         = {2013},
	journal      = pr,
	volume       = {46},
	number       = {6},
	pages        = {1658--1669}
}

@inproceedings{pham2014dropout,
	title        = {Dropout improves recurrent neural networks for handwriting recognition},
	author       = {Pham, Vu and Bluche, Th{\'e}odore and Kermorvant, Christopher and Louradour, J{\'e}r{\^o}me},
	year         = {2014},
	booktitle    = icfhr
}

@inproceedings{sanchez2014icfhr2014,
	title        = {{ICFHR2014 competition on handwritten text recognition on transcriptorium datasets (HTRtS)}},
	author       = {S{\'a}nchez, Joan Andreu and Romero, Ver{\'o}nica and Toselli, Alejandro H and Vidal, Enrique},
	year         = {2014},
	booktitle    = icfhr
}

@inproceedings{toselli2015handwritten,
	title        = {{Handwritten text recognition results on the Bentham collection with improved classical n-gram-HMM methods}},
	author       = {Toselli, Alejandro H and Vidal, Enrique},
	year         = {2015},
	booktitle    = hip
}

@inproceedings{jaderberg2015spatial,
	title        = {Spatial transformer networks},
	author       = {Jaderberg, Max and Simonyan, Karen and Zisserman, Andrew and Kavukcuoglu, Koray},
	year         = {2015},
	booktitle    = nips
}

@inproceedings{voigtlaender2016handwriting,
	title        = {Handwriting recognition with large multidimensional long short-term memory recurrent neural networks},
	author       = {Voigtlaender, Paul and Doetsch, Patrick and Ney, Hermann},
	year         = {2016},
	booktitle    = icfhr
}

@article{shi2016end,
	title        = {An end-to-end trainable neural network for image-based sequence recognition and its application to scene text recognition},
	author       = {Shi, Baoguang and Bai, Xiang and Yao, Cong},
	year         = {2016},
	journal      = ieeetpami,
	volume       = {39},
	number       = {11},
	pages        = {2298--2304}
}

@inproceedings{bluche2016joint,
	title        = {Joint line segmentation and transcription for end-to-end handwritten paragraph recognition},
	author       = {Bluche, Th{\'e}odore},
	year         = {2016},
	booktitle    = nips
}

@inproceedings{sanchez2016icfhr2016,
	title        = {{ICFHR2016 competition on handwritten text recognition on the READ dataset}},
	author       = {Sanchez, Joan Andreu and Romero, Veronica and Toselli, Alejandro H and Vidal, Enrique},
	year         = {2016},
	booktitle    = icfhr
}

@inproceedings{heusel2017gans,
	title        = {{GANs Trained by a Two Time-Scale Update Rule Converge to a Local Nash Equilibrium}},
	author       = {Heusel, Martin and Ramsauer, Hubert and Unterthiner, Thomas and Nessler, Bernhard and Hochreiter, Sepp},
	year         = {2017},
	booktitle    = nips
}

@inproceedings{puigcerver2017multidimensional,
	title        = {Are multidimensional recurrent layers really necessary for handwritten text recognition?},
	author       = {Puigcerver, Joan},
	year         = {2017},
	booktitle    = icdar
}

@inproceedings{bluche2017scan,
	title        = {{Scan, Attend and Read: End-to-End Handwritten Paragraph Recognition with MDLSTM Attention}},
	author       = {Bluche, Th{\'e}odore and Louradour, J{\'e}r{\^o}ome and Messina, Ronaldo},
	year         = {2017},
	booktitle    = icdar
}

@inproceedings{vaswani2017attention,
	title        = {Attention is all you need},
	author       = {Vaswani, Ashish and Shazeer, Noam and Parmar, Niki and Uszkoreit, Jakob and Jones, Llion and Gomez, Aidan N and Kaiser, {\L}ukasz and Polosukhin, Illia},
	year         = {2017},
	booktitle    = nips
}

@inproceedings{moysset2017full,
	title        = {Full-page text recognition: Learning where to start and when to stop},
	author       = {Moysset, Bastien and Kermorvant, Christopher and Wolf, Christian},
	year         = {2017},
	booktitle    = icdar
}

@inproceedings{bluche2017gated,
	title        = {Gated convolutional recurrent neural networks for multilingual handwriting recognition},
	author       = {Bluche, Th{\'e}odore and Messina, Ronaldo},
	year         = {2017},
	booktitle    = icdar
}

@article{sueiras2018offline,
	title        = {Offline continuous handwriting recognition using sequence to sequence neural networks},
	author       = {Sueiras, Jorge and Ruiz, Victoria and Sanchez, Angel and Velez, Jose F},
	year         = {2018},
	journal      = {Neurocomputing},
	volume       = {289},
	pages        = {119--128}
}

@inproceedings{such2018fully,
	title        = {Fully convolutional networks for handwriting recognition},
	author       = {Such, Felipe Petroski and Peri, Dheeraj and Brockler, Frank and Paul, Hutkowski and Ptucha, Raymond},
	year         = {2018},
	booktitle    = icfhr
}

@inproceedings{wigington2018start,
	title        = {{Start, Follow, Read: End-to-End Full-Page Handwriting Recognition}},
	author       = {Wigington, Curtis and Tensmeyer, Chris and Davis, Brian and Barrett, William and Price, Brian and Cohen, Scott},
	year         = {2018},
	booktitle    = eccv
}

@inproceedings{chowdhury2018efficient,
	title        = {An efficient end-to-end neural model for handwritten text recognition},
	author       = {Chowdhury, Arindam and Vig, Lovekesh},
	year         = {2018},
	booktitle    = bmvc
}

@inproceedings{strauss2018icfhr2018,
	title        = {{ICFHR2018 competition on automated text recognition on a READ dataset}},
	author       = {Strau{\ss}, Tobias and Leifert, Gundram and Labahn, Roger and Hodel, Tobias and M{\"u}hlberger, G{\"u}nter},
	year         = {2018},
	booktitle    = icfhr
}

@inproceedings{bhunia2019handwriting,
	title        = {{Handwriting Recognition in Low-Resource Scripts Using Adversarial Learning}},
	author       = {Bhunia, Ayan Kumar and Das, Abhirup and Bhunia, Ankan Kumar and Kishore, Perla Sai Raj and Roy, Partha Pratim},
	year         = {2019},
	booktitle    = cvpr
}

@article{moysset20192d,
	title        = {{Are 2D-LSTM really dead for offline text recognition?}},
	author       = {Moysset, Bastien and Messina, Ronaldo},
	year         = {2019},
	journal      = ijdar,
	volume       = {22},
	number       = {3},
	pages        = {193--208}
}

@inproceedings{zhang2019sequence,
	title        = {{Sequence-to-sequence domain adaptation network for robust text image recognition}},
	author       = {Zhang, Yaping and Nie, Shuai and Liu, Wenju and Xu, Xing and Zhang, Dongxiang and Shen, Heng Tao},
	year         = {2019},
	booktitle    = cvpr
}

@inproceedings{de2019no,
	title        = {{No padding please: Efficient neural handwriting recognition}},
	author       = {de Buy Wenniger, Gideon Maillette and Schomaker, Lambert and Way, Andy},
	year         = {2019},
	booktitle    = icdar
}

@inproceedings{michael2019evaluating,
	title        = {Evaluating sequence-to-sequence models for handwritten text recognition},
	author       = {Michael, Johannes and Labahn, Roger and Gr{\"u}ning, Tobias and Z{\"o}llner, Jochen},
	year         = {2019},
	booktitle    = icdar
}

@inproceedings{binkowski2018demystifying,
    title={{Demystifying {MMD} {GAN}s}},
    author={Mikołaj Bińkowski and Dougal J. Sutherland and Michael Arbel and Arthur Gretton},
    booktitle=iclr,
    year={2018}
}

@inproceedings{pippi2023hwd,
  title={{HWD: A Novel Evaluation Score for Styled Handwritten Text Generation}},
  author={Pippi, Vittorio and Quattrini, Fabio and Cascianelli, Silvia and Cucchiara, Rita},
  booktitle=bmvc,
  year={2023}
}

@article{cilia2019ranking,
	title        = {A ranking-based feature selection approach for handwritten character recognition},
	author       = {Cilia, Nicole Dalia and De Stefano, Claudio and Fontanella, Francesco and di Freca, Alessandra Scotto},
	year         = {2019},
	journal      = prl,
	volume       = {121},
	pages        = {77--86}
}

@inproceedings{clanuwat2019kuronet,
	title        = {{KuroNet: Pre-Modern Japanese Kuzushiji Character Recognition with Deep Learning}},
	author       = {Clanuwat, Tarin and Lamb, Alex and Kitamoto, Asanobu},
	year         = {2019},
	booktitle    = icdar
}

@inproceedings{baro2019towards,
  title={Towards a generic unsupervised method for transcription of encoded manuscripts},
  author={Bar{\'o}, Arnau and Chen, Jialuo and Forn{\'e}s, Alicia and Megyesi, Be{\'a}ta},
  booktitle={DATeCH},
  year = {2019}
}

@inproceedings{kang2020ganwriting,
	title        = {{GANwriting: Content-Conditioned Generation of Styled Handwritten Word Images}},
	author       = {Kang, Lei and Riba, Pau and Wang, Yaxing and Rusi{\~n}ol, Mar{\c{c}}al and Forn{\'e}s, Alicia and Villegas, Mauricio},
	year         = {2020},
	booktitle    = eccv
}

@inproceedings{yousef2020origaminet,
	title        = {{OrigamiNet: Weakly-Supervised, Segmentation-Free, One-Step, Full Page Text Recognition by learning to unfold}},
	author       = {Yousef, Mohamed and Bishop, Tom E},
	year         = {2020},
	booktitle    = cvpr
}

@inproceedings{cojocaru2020watch,
	title        = {Watch Your Strokes: Improving Handwritten Text Recognition with Deformable Convolutions},
	author       = {Cojocaru, Iulian and Cascianelli, Silvia and Baraldi, Lorenzo and Corsini, Massimiliano and Cucchiara, Rita},
	year         = {2020},
	booktitle    = icpr
}

@article{alkendi2024advancements,
  title={Advancements and challenges in handwritten text recognition: A comprehensive survey},
  author={AlKendi, Wissam and Gechter, Franck and Heyberger, Laurent and Guyeux, Christophe},
  journal={J. Imaging},
  year={2024},
}

@article{kang2022pay,
  title={Pay attention to what you read: non-recurrent handwritten text-line recognition},
  author={Kang, Lei and Riba, Pau and Rusi{\~n}ol, Mar{\c{c}}al and Forn{\'e}s, Alicia and Villegas, Mauricio},
  journal=pr,
  year={2022}
}

@inproceedings{aberdam2021sequence,
	title        = {{Sequence-to-Sequence Contrastive Learning for Text Recognition}},
	author       = {Aberdam, Aviad and Litman, Ron and Tsiper, Shahar and Anschel, Oron and Slossberg, Ron and Mazor, Shai and Manmatha, R and Perona, Pietro},
	year         = {2021},
	booktitle    = cvpr,
}

@article{li2021trocr,
	title        = {{TrOCR: Transformer-based optical character recognition with pre-trained models}},
	author       = {Li, Minghao and Lv, Tengchao and Cui, Lei and Lu, Yijuan and Florencio, Dinei and Zhang, Cha and Li, Zhoujun and Wei, Furu},
	year         = {2023},
	journal      = aaai
}

@article{wick2021rescoring,
	title        = {{Rescoring Sequence-to-Sequence Models for Text Line Recognition with CTC-Prefixes}},
	author       = {Wick, Christoph and Z{\"o}llner, Jochen and Gr{\"u}ning, Tobias},
	year         = {2021},
	journal      = {arXiv preprint arXiv:2110.05909}
}

@inproceedings{cascianelli2021learning,
	title        = {{Learning to Read L'Infinito: Handwritten Text Recognition with Synthetic Training Data}},
	author       = {Cascianelli, Silvia and Cornia, Marcella and Baraldi, Lorenzo and Piazzi, Maria Ludovica and Schiuma, Rosiana and Cucchiara, Rita},
	year         = {2021},
	booktitle    = icpr
}

@inproceedings{cascianelli2022lam,
	title        = {{The LAM Dataset: A Novel Benchmark for Line-Level Handwritten Text Recognition}},
	author       = {Cascianelli, Silvia and Pippi, Vittorio and Martin, Maarand and Cornia, Marcella and Baraldi, Lorenzo and Christopher, Kermorvant and Cucchiara, Rita},
	year         = {2022},
	booktitle    = icpr
}

@article{yang2023gpt4v,
	title        = {{The Dawn of LMMs: Preliminary Explorations with GPT-4V(ision)}},
	author       = {Yang, Zhengyuan and Li, Linjie and Lin, Kevin and Wang, Jianfeng and Lin, Chung-Ching and Liu, Zicheng and Wang, Lijuan},
	year         = {2023},
	journal      = {arXiv preprint arXiv:2309.17421}
}

@inproceedings{baena2024general,
	title        = {General Detection-based Text Line Recognition},
	author       = {Baena, Raphael and Kalleli, Syrine and Aubry, Mathieu},
	year         = {2024},
	booktitle    = nips
}

@inproceedings{arce2022self,
  title={Self-attention networks for non-recurrent handwritten text recognition},
  author={d’Arce, Rafael and Norton, Terence and Hannuna, Sion and Cristianini, Nello},
  booktitle= icfhr,
  year={2022}
}

@article{li_htr-vt_2025,
	title = {{HTR}-{VT}: Handwritten Text Recognition with Vision Transformer},
	author = {Li, Yuting and Chen, Dexiong and Tang, Tinglong and Shen, Xi},
    journal = pr,
    year = 2025
}

@article{training_barrere_2024,
	title        = {Training transformer architectures on few annotated data: an application to historical handwritten text recognition},
	author       = {Barrere, Killian and Soullard, Yann and Lemaitre, Aurélie and Coüasnon, Bertrand},
	year         = 2024,
	journal      = ijdar
}

@inproceedings{light_barrere_2022,
	title        = {A Light Transformer-Based Architecture for Handwritten Text Recognition},
	author       = {Barrere, Killian and Soullard,  Yann  and Lemaitre, Aurélie  and Coüasnon, Bertrand},
    booktitle    = {DAS},
	year         = 2022
}

@article{coquenet2023endtoend,
   title={End-to-End Handwritten Paragraph Text Recognition Using a Vertical Attention Network},
   journal=ieeetpami,
   author={Coquenet, Denis and Chatelain, Clement and Paquet, Thierry},
   year={2023},
}

@inproceedings{wolf-etal-2020-transformers,
    title = "Transformers: State-of-the-Art Natural Language Processing",
    author = "Thomas Wolf and Lysandre Debut and Victor Sanh and Julien Chaumond and Clement Delangue and Anthony Moi and Pierric Cistac and Tim Rault and Rémi Louf and Morgan Funtowicz and Joe Davison and Sam Shleifer and Patrick von Platen and Clara Ma and Yacine Jernite and Julien Plu and Canwen Xu and Teven Le Scao and Sylvain Gugger and Mariama Drame and Quentin Lhoest and Alexander M. Rush",
    booktitle = emnlp,
    year = "2020",
}

@article{otsu1979threshold,
  title   = {A threshold selection method from gray-level histograms},
  author  = {Otsu, Nobuyuki},
  journal = ieeetsmc,
  year    = {1979}
}

@inproceedings{quattrini2024fourbi,
  title        = {{Binarizing Documents by Leveraging Both Space and Frequency}},
  author       = {Quattrini, Fabio and Pippi, Vittorio and Cascianelli, Silvia and Cucchiara, Rita},
  booktitle    = icdar,
  year         = {2024},
}

@article{gillelevenson2023castilian,
  title        = {Towards a General Open Dataset and Models for Late Medieval Castilian Text Recognition (HTR/OCR)},
  author       = {Gille Levenson, Matthias},
  journal      = {J. Data Min. Digit. Humanit.},
  year         = {2023}
}

@article{miyagawa2019coptic,
  title        = {Optical Character Recognition of Typeset Coptic Text with Neural Networks},
  author       = {Miyagawa, So and Bulert, Kirill and B{\"u}chler, Marco and Behlmer, Heike},
  journal      = {Digit. Scholarsh. Humanit},
  volume       = {34},
  number       = {Supplement 1},
  pages        = {i135--i141},
  year         = {2019}
}

@inproceedings{lincke2019coptic,
  title        = {Optical Character Recognition for Coptic Fonts: A Multi-Source Approach for Scholarly Editions},
  author       = {Lincke, Eliese-Sophia and Bulert, Kirill and B{\"u}chler, Marco},
  booktitle    = {DATeCH},
  year         = {2019}
}

@misc{coptot,
  title        = {{CoptOT}: Digital Edition of the Coptic Old Testament},
  author       = {{CoptOT}},
  howpublished = {\url{https://coptot.manuscriptroom.com}}
}

@misc{coptic_scriptorium,
  title        = {{Coptic Scriptorium}: Digital Research in Coptic Language and Literature},
  author       = {{Coptic Scriptorium}},
  howpublished = {\url{https://copticscriptorium.org}}
}

\end{document}